\documentclass[letterpaper]{article} % DO NOT CHANGE THIS
\usepackage{aaai24}  % DO NOT CHANGE THIS
\usepackage{times}  % DO NOT CHANGE THIS
\usepackage{helvet}  % DO NOT CHANGE THIS
\usepackage{courier}  % DO NOT CHANGE THIS
\usepackage[hyphens]{url}  % DO NOT CHANGE THIS
\usepackage{graphicx} % DO NOT CHANGE THIS
\usepackage{natbib}  % DO NOT CHANGE THIS AND DO NOT ADD ANY OPTIONS TO IT
\usepackage{caption} % DO NOT CHANGE THIS AND DO NOT ADD ANY OPTIONS TO IT
\usepackage{algorithm}
\usepackage{algorithmic}
\usepackage{multirow} 
\usepackage{soul}
\usepackage{url}
\usepackage{anyfontsize}
\usepackage{booktabs}
\usepackage{algorithm}
\usepackage{algorithmic}
\usepackage[switch]{lineno}
\usepackage{amsfonts}
\usepackage{amssymb}
\usepackage{indentfirst}
\usepackage{bbding}
\usepackage{colortbl}
\usepackage{amsmath}
\usepackage[table]{xcolor}

\usepackage{newfloat}
\usepackage{listings}

\usepackage{hyperref}

\DeclareCaptionStyle{ruled}{labelfont=normalfont,labelsep=colon,strut=off} % DO NOT CHANGE THIS
\floatstyle{ruled}
\newfloat{listing}{tb}{lst}{}
\floatname{listing}{Listing}
\title{Less is More: Learning Reference Knowledge Using No-Reference \\ Image Quality Assessment}
\author{
    Xudong Li\textsuperscript{\rm 1},
    Jingyuan Zheng\textsuperscript{\rm 2},
    Xiawu Zheng\textsuperscript{\rm 3},
    Runze Hu\textsuperscript{\rm 4},
    Enwei Zhang\textsuperscript{\rm 5},
    Yuting Gao\textsuperscript{\rm 5},\\
    Yunhang Shen\textsuperscript{\rm 5},
    Ke Li\textsuperscript{\rm 5},
    Yutao Liu\textsuperscript{\rm 6},
    Pingyang Dai\textsuperscript{\rm 1},
    Yan Zhang\textsuperscript{\rm 1,†},
    Rongrong Ji\textsuperscript{\rm 1}
}
\affiliations{
    \textsuperscript{\rm 1}Key Laboratory of Multimedia Trusted Perception and Efficient Computing,\\ Ministry of Education of China, Xiamen University\\
    \textsuperscript{\rm 2} School of Medicine, Xiamen University
    \textsuperscript{\rm 3} Peng Cheng Laboratory, Shenzhen\\
    \textsuperscript{\rm 4} School of Information and Electronics, Beijing Institute of Technology\\
    \textsuperscript{\rm 5} Tencent Youtu Lab
    \textsuperscript{\rm 6} School of Computer Science and
Technology, Ocean University of China\\
\{lxd761050753, jyzheng0606,
bzhy986, hrzlpk2015, shenyunhang01, tristanli.sh\}@gmail.com,\\
zhengxw01@pcl.ac.cn,
\{miyozhang, yutinggao\}@tencent.com,
liuyutao@ouc.edu.cn,\{pydai, rrji\}@xmu.edu.cn

}

\usepackage{bibentry}
\begin{document}
\maketitle
\begin{abstract}
    \emph{Image Quality Assessment} (IQA) with reference images have achieved great success by imitating the human vision system, in which the image quality is effectively assessed by comparing the query image with its pristine reference image. 
    However, for the images in the wild, it is quite difficult to access accurate reference images. 
    We argue that it is possible to learn reference knowledge under the \emph{No-Reference Image Quality Assessment} (NR-IQA) setting, which is effective and efficient empirically.
    Concretely, by innovatively introducing a novel feature distillation method in IQA, we propose a new framework to learn comparative knowledge from non-aligned reference images. 
    And then, to achieve fast convergence and avoid overfitting, we further propose an inductive bias regularization.    
    Such a framework not only solves the congenital defects of NR-IQA but also improves the feature extraction framework, enabling it to express more abundant quality information.
    Surprisingly, our method utilizes \textbf{less} input while obtaining a \textbf{more} significant improvement compared to the teacher models.
    Extensive experiments on $8$ standard NR-IQA datasets demonstrate the superior performance to the state-of-the-art NR-IQA methods, i.e., achieving the PLCC values of 0.917 (\textcolor{red}{$\uparrow 3.3\%$} vs. 0.884 in LIVEC) and 0.686 (\textcolor{red}{$\uparrow 2.5\%$} vs. 0.661 in LIVEFB).
\end{abstract}
\begin{figure}[t]
  \centering
    \includegraphics[width=0.5\textwidth]{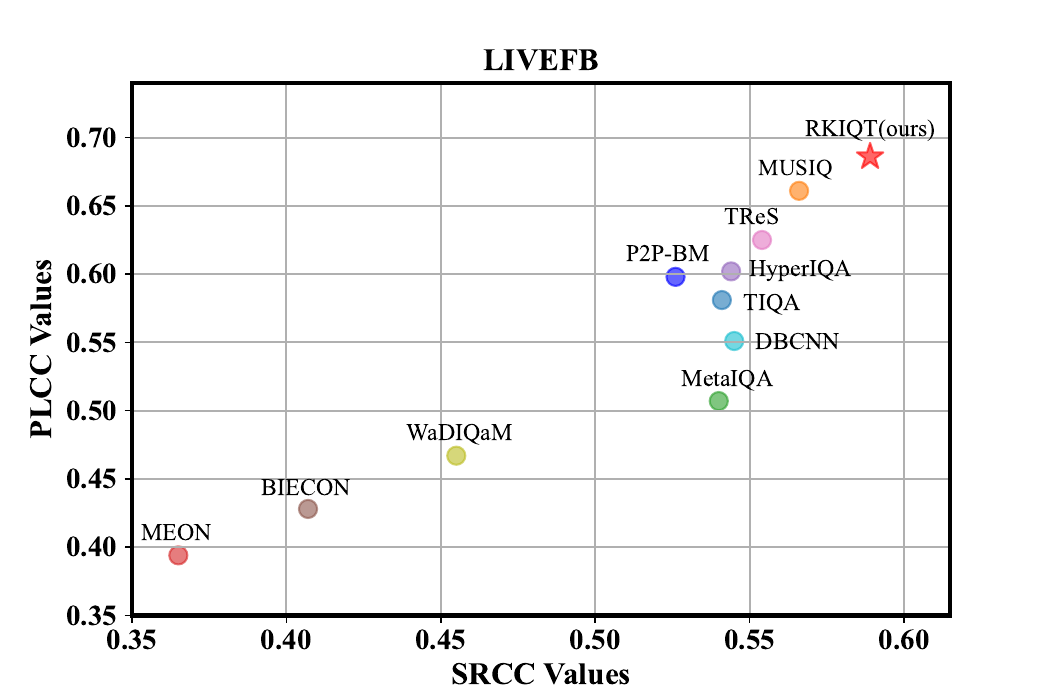}
  \caption{The performance comparison between the proposed RKIQT and different NR-IQA methods on LIVEFB (largest authentic dataset). The proposed RKIQT outperforms all existing NR-IQA methods. Note that it also exceeds some of SOTA \textbf{Full-Reference} IQA methods (DeepIQA~\cite{deepQA} and DISTS \cite{DISTS}).}
  \label{fig1}
\end{figure}

\section{Introduction}
    \emph{Image Quality Assessment} (IQA)~\cite{end,rehman2012reduced,gu2017learning,qpt,reiqa} has been applied in many computer vision pieces of research including but not limited to image restoration~\cite{banham1997digital} and super-resolution~\cite{dong2015image}.
    By imitating the human vision system, IQA methods effectively estimate the quality of the query image with its pristine reference image, which achieves promising results with massive data support.
    For instance, a previous study~\cite{wang2004image} based on hand-crafted proposed a structural similarity index method that uses all or some of the information from \emph{High Quality} (HQ) reference images to evaluate image quality, making substantial progress toward establishing full reference image quality.
    Learning-based method~\cite{cheon2021perceptual} utilizes different information by comparing pixel-aligned HQ reference and distorted images to create a more accurate and consistent assessment of the quality of distorted images.
    
    However, the pristine reference image is not always available in practice.
    Some researchers propose to leverage the texture information with a supervision framework to infer the quality of the query image and propose \emph{No-Reference IQA} (NR-IQA) methods~\cite{fang2020perceptual,li2018has,bosse2017deep, LIQE, DEIQT,zhang2021uncertainty}.
    For example, Zhang \emph{et al.}~\cite{zhang2018blind} uses pre-trained CNN modules~\cite{simonyan2014very} for fine-tuning to assess image quality which achieves better generalizability, and robustness.
    With the popularity of ViT ~\cite{ViT} in computer vision, state-of-the-art (SOTA) NR-IQA methods employ ViT-based architectures~\cite{dosovitskiy2020image}, which perform an end-to-end optimization of feature engineering and quality regression, simultaneously.
    Although such a pipeline~\cite{zhu2022blind} solves the problem of missing reference images, its performance is far from satisfactory.
    Here we provide a biological explanation, which is also known as consensus in psychology.
    The human vision system is easier to perceive image quality degradation by comparing two images rather than a single sample.%噪声第14章第5节“匹配中的噪声：绝对判断的局限性”
    Therefore, NR-IQA methods that give up the comparison knowledge naturally degrade the performance of the algorithm~\cite{ponomarenko2009tid2008,liang2016image,yin2022content}.

    In this paper, we propose a novel NR-IQA framework, herein namely \emph{Reference Knowledge-guided image quality transformer} (RKIQT) to perform the inference procedure of IQA with the reference information learned during training.
    Thus, a novel {Masked Quality-Contrastive Distillation} (MCD) is introduced as a feature distillation method to obtain comparison knowledge.
    To achieve fast convergence and avoid overfitting, we further propose an inductive bias regularization.
%    the proposed method advances in two aspects: (\romannumeral1) The reference image is only used in training stage, which ensures the effectiveness of our proposed method. (\romannumeral2) The teachers are settled in different stages, which ensures their independence and allows them to maximize their effectiveness for the whole framework. Our contributions are listed as follows:
    \begin{itemize}
        \item We creatively leverage the feature distillation to achieve the comparison knowledge under the NR-IQA setting. Since the teacher is a non-aligned IQA network, there is no need for a pristine high-quality reference image.
        \item For feature distillation, we introduce a Masked Quality-Contrastive Distillation to guide students in imitating the teacher's prior comparison information based on partial feature pixels, which implies our students will be more robust and have stronger representation capacity.
        \item For regularization, we leverage the reverse distillation strategy while distilling teachers and tokens with different inductive biases. while speeding up the training process, we adapt students to this reverse distillation to obtain more competitive quality-aware benefits by fine-tuning the quality-aware ability of pre-trained teachers.
     \end{itemize}
    
    Extensive experiments demonstrate the effectiveness and efficiency of our method. In particular, we verify RKIQT on 8 benchmark IQA datasets involving a wide range of image contents, distortion types, and dataset size. RKIQT outperforms other competitors across all these datasets.
    
\section{Related Work}
\textbf{NR-IQA with Deep Learning}. The deep learning methods have achieved extraordinary success in various computer vision tasks, which by nature attracts a great deal of interest in utilizing deep learning for IQA tasks. 
The early version of deep learning-based IQA method~\cite{ Liu_2017_ICCV,zhang2018blind, hypernet,zhang2022continual} is based on the convolutional neural network (CNN)~\cite{he2016deep} thanks to its powerful feature expression ability. The CNN-based IQA method generally treats the IQA task as the downstream task of object recognition, following the standard pipeline of pre-training and fine-tuning. 
Such a strategy is useful as these pre-trained features share a certain degree of similarity with the quality-aware features of images~\cite{hypernet}. 
Recently, the Vision Transformer (ViT)~\cite{ViT} based NR-IQA methods are growing in popularity, owing to the strong capability of ViT in modeling the non-local perceptual features of the image. There are mainly two types of architectures for the ViT-based NR-IQA methods, including hybrid Transformer~\cite{TReS} and pure ViT-based Transformer~\cite{musiq}. 
The hybrid architecture generally combines the CNNs with the Transformer, which are responsible for the local and long-range feature characterization, respectively.  
The ViT-based methods can be further exploited. Existing ViT methods generally rely on the Class~(CLS) token to judge the image quality. The CLS token is initially designed to describe the image content (object recognition), and thus the preserved features are mainly related to the higher-level visual abstractions, i.e., semantics and spatial relationships of objects.
As a result, using the CLS token solely is not adequate in characterizing the quality-aware features of an image. 
\begin{figure*}[htbp]
	\centering{\includegraphics[width=177mm]{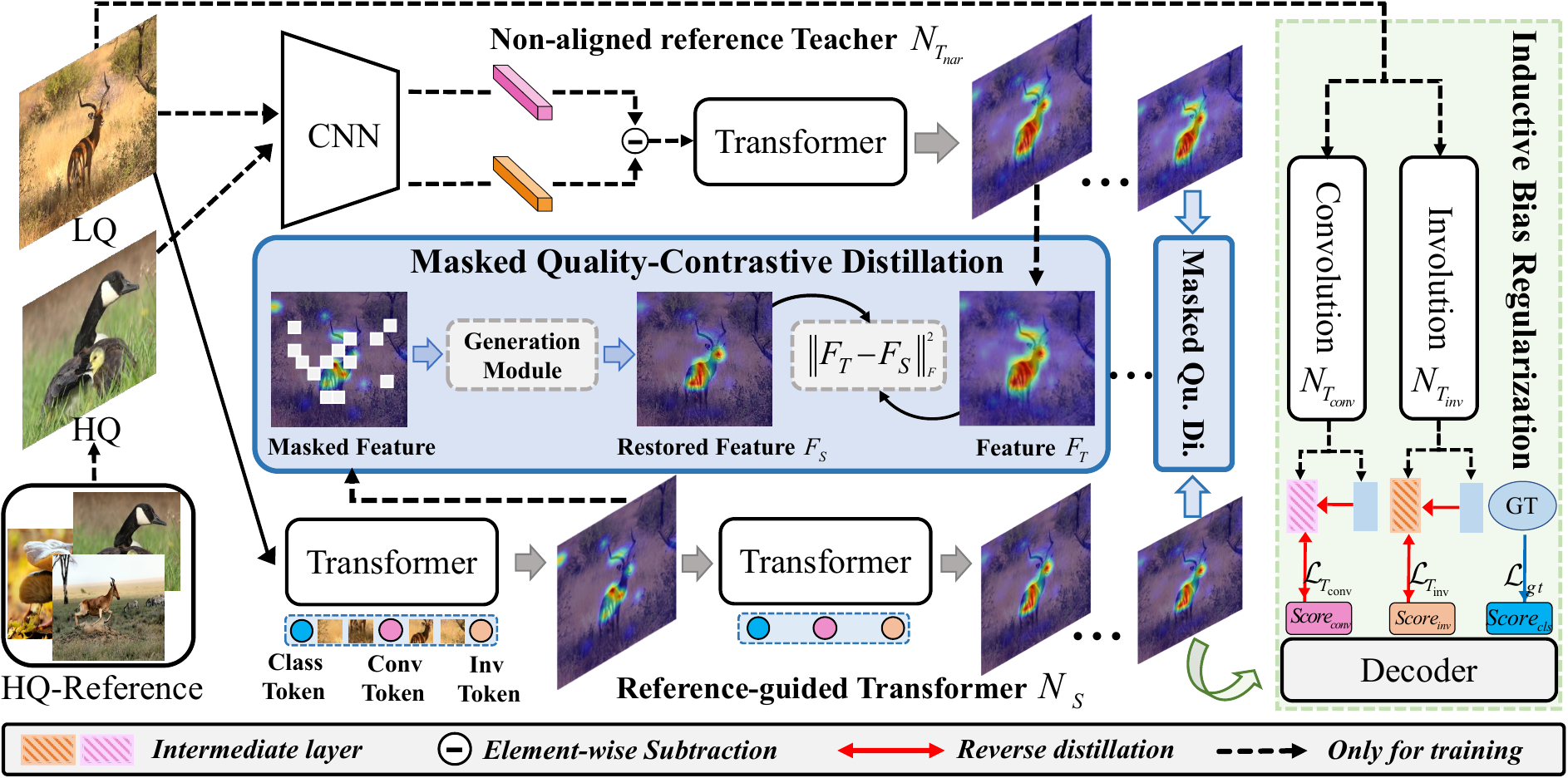}}
	\caption{The overview of our RKIQT. We first mask the feature map of the student network, which is then used to generate the new feature that is supervised by a Non-aligned reference teacher network (Sec.~\ref{sec:mcd}). 
 After that, we further propose inductive bias regularization, which extracts local and global knowledge from CNN and Involution respectively to achieve fast convergence and avoid overfitting (~\ref{sec:regular}).
 }
   \label{modelStructure}
 \vspace{-10pt}
\end{figure*}
\section{The Proposed Method}
To make it clear, we use bold format denotes as vectors (\emph{e.g.,} $\boldsymbol{x, y}$), matrices (\emph{e.g.,} $\boldsymbol{X, Y}$) or tensors. We further define some common notations in IQA. 
In particular, we define the Low Quality (LQ) image to be estimated as $I_L$, 
the randomly selected annotated High-Quality (HQ) image as $I_H$,
the feature map of the network output as $\boldsymbol F$,
the quality prediction of network $\boldsymbol N$ is denoted as $Y$.

IQA is highly correlated to subjective cognition, which is more accurate when the pristine reference image is provided~\cite{wang2004image}. 
However, it is impractical to find reference images in real-world applications. In this paper, we propose a novel framework that learns reference information under the NR-IQA setting. It consists of three dedicatedly designed components: (i) The NR-student Reference Knowledge-guided Image Quality Transformer (RKIQT) {$\boldsymbol N_{s}$} is the main network of our method, which receives the knowledge from other teacher networks. (ii) The non-aligned reference teacher~(NAR-teacher) {$\boldsymbol N_{{T_{nar}}}$} offers the comparison knowledge to {$\boldsymbol N_{s}$} by Masked Quality-Contrastive Distillation. (iii) The inductive bias teachers {$\boldsymbol N_{{T_{conv}}}$}, {$\boldsymbol N_{{T_{inv}}}$} provide the quality-aware knowledge to {$\boldsymbol N_s$} by Inductive Bias Regularization.

As illustrated in Fig.~\ref{modelStructure}, given input images, our student and NAR-teacher first obtain the LQ local-global fused features and the HQ-LQ distribution difference through the outputs of the transformer encoder, respectively. The student's feature map is first masked and then used to reconstruct a new feature through a simple generation module, which is supervised by the teacher (Sec.~\ref{sec:mcd}). Then, we further propose inductive bias regularization, which extracts local and global knowledge from CNN and Involution ~\cite{li2021involution} respectively to achieve fast convergence and avoid overfitting (Sec.~\ref{sec:regular}). After training, all teacher models and regularization will be deprecated, the student model is directly applied for inference without reference image.
\subsection{Masked Quality-Contrastive Distillation}~\label{sec:mcd}
We make the first attempt to transfer HQ-LQ differential prior information from non-aligned reference teacher~(Sec.~\ref{sec:archi}) to NR-IQA via Knowledge Distillation (KD).
Traditional KD methods require the student model to directly mimic the teacher model's output. 
Such a mechanism is not suitable for our method, since our student model lacks reference images, it can only mine the quality features of LQ images. It is misaligned with the HQ-LQ distribution difference features captured by the teacher. Direct imitation teacher's output may introduce negative regularization that degrades the final performance and stability~(Sec.~\ref{ablation_mcd}).  

Inspired by the recent effective application of the masking mechanism \cite{mae,mgd}, this paper proposes a simple yet effective feature distillation method, named Masked Quality Contrastive Distillation (MCD). The goal of the proposed MCD is not to directly mimic the HQ-LQ difference features extracted by the teacher but rather to use these features to guide students in developing an awareness for making comparisons. Specifically, we first randomly mask the student features and then force the student model to generate the teacher's complete features using a simple feature generation module.
The improvement in the student model's comparison skills manifests in two key aspects. First, the student model restores the teacher's features through the per-patch features of the mask, which enhances the awareness to contrast locally~\cite{mae}. Secondly, the student's skill in extracting information about global edge contours for contrast is enhanced, guided by the teacher‘s differences in HQ-LQ prior distributions~(Sec.~\ref{sec:attentionmap}).
% During the MCD process in each layer, the student model employs random pixels, ensuring that every pixel is utilized throughout the whole training.  This approach leads to the development othe extractionf more robust features with enhanced representational capabilities. 

More specifically, given the ${{i}_{th}}$ image, all inner features $\boldsymbol F_\text{$T$}^{(i)}$ of the NAR-teacher are applied to guide the training of NR-student. We first set the random mask function $ {M(\cdot)}$ to cover the corresponding student's feature which has passed through an adaptive layer to align the teacher feature map. Then, the student’s feature is covered with the corresponding mask, which is used to generate new feature maps through a generation module ${\mathcal{G(\cdot)}}$ includes two 3×3 convolutional layers with ReLU. Finally, MSE loss regression is used as the feature distillation loss to transfer knowledge to the image's corresponding layer features $\boldsymbol F_\text{$S$}^{(i)}$ of NR-student, which is expressed as follows:
\begin{equation}
\begin{aligned}
    \boldsymbol F_\text{$S$}^{(i)}&= {\mathcal{G}(M( {\boldsymbol F_{{S}'}^{(i)}   }))}\\
    {\mathcal{L}_{\text{feature}}}(\boldsymbol F_S,\boldsymbol F_T) &=\frac{1}{K} \sum\limits_{{i=1}}^{K} \|{\boldsymbol 
F_\text{$T$}^{(i)} -\boldsymbol F_\text{$S$}^{(i)}}  \|_{F}^2
\end{aligned}
\end{equation}
\noindent where ${\boldsymbol F_{{S}'}^{(i)}}$ represent the aligned feature map of the student encoder , $K$ denotes the number of images in the training set. Guided by MCD, our student effectively learns more HQ-LQ difference knowledge and remains stable across differently distorted images.
\subsection{Inductive Bias Regularization}~\label{sec:regular}
Prior research~\cite{wang2019distilling} found that the performance gains of KD mainly come from the regularization of the logits-based knowledge in the teacher model. Therefore, to achieve fast convergence and avoid overfitting, we further propose an inductive bias regularization that adopts EfficientNet-b0  ~\cite{tan2019efficientnet} and RedNet101~\cite{li2021involution} (pre-trained on ImageNet~\cite{imagenet}) which considers the trade-off between accuracy and complexity to guide the student to obtain more comprehensive representation power. \footnote{We do not use VIT ~\cite{ViT} due to its fewer inductive biases ~\cite{cross} which will make IQA focus on local features}. To explain, CNN has a strong locality modeling capability, while the involution kernel is shared across channels but distinct in the spatial extent, and dynamically generating kernel parameters, which enables the extraction of long-range spatial information in images. 

However, We believe that if teachers' logits with different inductive biases are directly used to supervise students, there will be a relatively large quality perception gap between teacher and student. Therefore, we introduce a learnable intermediate layer to solve such a problem. 
Specifically, the introduced learnable intermediate layer is proposed to learn the output of the corresponding teacher network and also takes the supervision information from the student network.
Take INN as an example~(same with CNN branches), given the ${{i}_{\text{th}}}$ image, the teacher’s output is defined as ${\boldsymbol Y_{{T}'_{inv}   }}$. 
Meanwhile, the output of the teacher's learnable intermediate layer and student network is defined as ${\boldsymbol Y_{{T}_{inv}}   }$ and $\boldsymbol Y_{{S}_{inv}  }$, respectively, which is expressed as follows:
\begin{equation}
        {\boldsymbol Y_{{T}_{inv}}   } =  \text{MLP}({( {{{ A}_1}( {{F_{{1}}}} ) \oplus {{ A}_2}( {{F_{{2}}}} )} ) \oplus {{ A}_3}( {{F_{{3}}}} )} )
\label{equ2}
\end{equation}
where $(F_{1}, F_{2}, F_{3})$ donates the feature of different middle layers of the pre-trained Teacher network, transformed through the feature adaptation layer $\boldsymbol A(\cdot)$ and feature addition $\oplus$. During training, The $\mathcal L_1$ regression is adopted as the distillation loss, and the loss function of the student and intermediate layer is mathematically expressed as:
\begin{equation}
   \small {\mathcal{L}_{{S}_{inv}}} = {\frac{1}{K} \sum\limits_{i = 1}^K {\|  {\boldsymbol Y_{{S}_{inv}}^{(i)}   } -{\boldsymbol Y_{{T}_{inv}}^{(i)}   } \| }_1}
\label{equ3}
\end{equation}
\begin{equation}
    \small {\mathcal{L}_{{T}_{inv}}} = {\frac{1}{K} \sum\limits_{i = 1}^K {\|  {\boldsymbol Y_{{T}_{inv}}^{(i)}   } -{\boldsymbol Y_{{T}'_{inv}}^{(i)}   } \| }_1}+\frac{1}{K} \sum\limits_{i = 1}^K \|{ {{ {\boldsymbol Y_{{T}_{inv}}^{(i)}   } - {\boldsymbol Y_{{S}_{inv}}^{(i)}   }}\|}}_1.
\label{equ4}
\end{equation}
\noindent In this way, the ability gap between teachers and students is effectively narrowed. 
Meanwhile, the students even outperform the teacher and get a noticeable improvement. From the perspective of a student, the output takes supervision from two teachers, which is formally defined as:
% \begin{equation}
%     {{ L}_{conv}} =  ( {{f_{{S_{conv}}}},{f_{{T^\prime }}}_{_{conv}}}  ).
% \end{equation}
% \begin{equation}
%     {{ L}_{inv}} =  ( {{f_{{{S}_{inv}}}},{f_{{T^\prime }_{inv}}}}  ).
% \end{equation}
\begin{equation}
    {\mathcal{L}_{\text{Logits}}} =  {{\mathcal{L}_{{S}_{inv}}}} + {{\mathcal{L}_{{S}_{conv}}}},
    \label{equ5}
\end{equation}
where the calculation process of ${\mathcal{L}_{{S}_{conv}}}$ is similar to ${\mathcal{L}_{{S}_{inv}}}$. Take the ground truth as extra supervision, the loss function of the student is finally formally defined as:
% \subsection{Overall loss function}
%我们对学生模型进行端到端的训练,蒸馏学生模型的总Loss如下:
% \begin{equation}
%     \mathcal{L} = \frac{1}{N}\sum\limits_{i = 1}^N {{{ \| {{y_i} - {N_s} ( {\boldsymbol I_\text{$L$}^{(i)};{\theta _s}}  )}  \|}_1}} + {\lambda _1}{\mathcal{L}_{{ \text{Feature }}}} + {\lambda _2}{\mathcal{L}_\text{Logits}},
% \end{equation}
\begin{equation}
    \mathcal{L} = \frac{1}{K}\sum\limits_{i = 1}^K {{{ \| {\boldsymbol Y_{gt}^{(i)} - \boldsymbol{N_s} (\boldsymbol I_\text{$L$}^{(i)} )}  \|}_1}} + {\lambda _1}{\mathcal{L}_{{\text{Feature }}}} + {\lambda _2}{\mathcal{L}_{\text{Logits}}},
    \label{loss equ}
\end{equation}
%其中L$_{GT}$为模型的$L1$回归损失,${\lambda_1、\lambda_2、\lambda_3}$为同一尺度下平衡各损失的超参数。
\noindent where $\boldsymbol I_\text{$L$}^{(i)}$ is the ${{i}_{\text{th}}}$ distorted image,${N_{s}}$(·) is the student predicted results and labeled ground-truth is represented as $\boldsymbol Y_{gt}^{(i)}$. $\lambda_1, \lambda_2$ are the hyperparameters.

\subsection{Architecture Design}~\label{sec:archi}
\noindent \textbf{Non-aligned reference Teacher Network.}
In the NR-IQA task, the main noise occurs in the fine-tuning process from the recognition task to the IQA task, since the recognition task has no direct relation to image quality.
To reduce the noise, we utilize a non-aligned IQA teacher~(NAR-teacher) to offer more reliable comparison knowledge during training.
We use an offline distillation scheme with the pre-trained Inception-ResNet-v2~\cite{szegedy2017inception} network to extract feature maps from the non-aligned reference and distorted input images for our NAR-teacher network, which has the same architecture as ~\cite{ViT} but with different inputs.  By comparing the differences between the non-aligned reference and input images, the NAR-teacher network offers valuable comparative knowledge to optimize the student network for the NR-IQA task.

% \noindent \textbf{Student.}
\noindent \textbf{Reference-guided Transformer Student}. 
The transformer encoder can comprehensively characterize an image's perceptual features by aggregating both local and global information \cite{dosovitskiy2020image}. However, transformers with fewer inductive biases may struggle when trained with limited data. This issue can be addressed through the distillation technique~\cite{zhu2018knowledge,gou2021knowledge,phuong2019towards}, where a student model with smaller inductive biases can learn various knowledge from teachers with different inductive biases~\cite{cross}. In this regard, we propose cross-inductive bias teachers that can focus on local quality degradation and global quality perception. To align tokens with different inductive biases, we introduce token inductive bias alignment. We use three tokens: Class token, Conv token, and Inv token. We apply truncated Gaussian initialization to the Class token to eliminate its inductive bias and align it with the ground truth~\cite{touvron2021training}. On the other hand, we introduce the corresponding inductive bias into the remaining two tokens. The Conv token and Inv token use the average pooling outputs of convolution stem~\cite{graham2021levit} and involution stem, respectively, with added position embeddings. The output of the encoder includes three inductive bias tokens denoted by $\hat{\boldsymbol F_o}\in \mathbb{R}^{3\times D}$.

Previous works~\cite{DEIQT} found that CLS tokens cannot build an optimal representation for image quality. To this end, we introduce a quality-aware decoder to further decode inductive biases CLS, Conv, and Inv tokens through multi-head self-attention (MHSA), thus making the extracted features more significant and comprehensive to the image quality. The queries $Q_d$ of the decoder are written by:
\begin{equation}
            {\boldsymbol Q_d} = {\text{ MHSA}}  ( {{\text{Norm}} ( {\hat{\boldsymbol F_o}} + \boldsymbol J)}) + ( {\hat{\boldsymbol F_o}}+ \boldsymbol J ),
\end{equation}%
\noindent where $J \in  \mathbb{R}^{3\times D}$ is initialized with random numbers,  which evaluate the image quality from different perspectives~\cite{DEIQT}.
% $P = \{P_{1}, P_{2}, P_{3}\} \in \mathbb{R}^{3\times D}$ represents the quality-aware features with different inductive bias tokens from the decoder output, which is more comprehensive and accurate in defining image quality. $L$ is calculated as in:
\begin{equation}
            {\boldsymbol{\hat Y}} = {\text{MLP}}  ( {{\text{MHCA}}  ( {{\text{Norm}}  ( {{{{\boldsymbol Q}}_d}}  ),{{{\boldsymbol K}}_d},{{{\boldsymbol V}}_d}}  ) + {{{\boldsymbol Q}}_d}}  )
\end{equation}%
During Multi-Head Cross-Attention~(MHCA), we utilize $Q_d$ to interact with the features of the image patches preserved in the encoder outputs. The results are then fed to an MLP to derive the final quality score ${\boldsymbol{\hat Y}}$. The quality-aware decoder can significantly improve the learning ability of the transformer-based NR-IQA model, thus improving the performance of the model and generalization ability.
\begin{center} 
\begin{table*}[t]
\setlength\tabcolsep{0.7pt}
    \centering
    \resizebox{1\textwidth}{!}{
        \begin{tabular}{lcccccccc||cccccccc}
      \toprule[1.5pt]
    & \multicolumn{2}{c}{LIVE} & \multicolumn{2}{c}{CSIQ} & \multicolumn{2}{c}{TID2013} & \multicolumn{2}{c||}{KADID} & \multicolumn{2}{c}{LIVEC} & \multicolumn{2}{c}{KonIQ} & \multicolumn{2}{c}{LIVEFB} & \multicolumn{2}{c}{SPAQ}\\
    \cmidrule{2-17}    Method & \multicolumn{1}{c}{PLCC} & \multicolumn{1}{c}{SRCC} & \multicolumn{1}{c}{PLCC} & \multicolumn{1}{c}{SRCC} & \multicolumn{1}{c}{PLCC} & \multicolumn{1}{c}{SRCC}& \multicolumn{1}{c}{PLCC} & \multicolumn{1}{c||}{SRCC}& \multicolumn{1}{c}{PLCC} & \multicolumn{1}{c}{SRCC}& \multicolumn{1}{c}{PLCC} & \multicolumn{1}{c}{SRCC}& \multicolumn{1}{c}{PLCC} & \multicolumn{1}{c}{SRCC}& \multicolumn{1}{c}{PLCC} & \multicolumn{1}{c}{SRCC}\\
    \midrule
    DIIVINE ~\cite{saad2012blind} & 0.908 & 0.892 & 0.776 & 0.804 & 0.567 & 0.643 & 0.435 & 0.413 & 0.591 & 0.588 & 0.558 & 0.546 & 0.187 & 0.092 & 0.600 & 0.599 \\
    BRISQUE ~\cite{BRISQUE} & 0.944 & 0.929 & 0.748 & 0.812 & 0.571 & 0.626 & 0.567 & 0.528 & 0.629 & 0.629 & 0.685 & 0.681 & 0.341 & 0.303 & 0.817 & 0.809 \\
    ILNIQE ~\cite{ILNIQE} & 0.906 & 0.902 & 0.865 & 0.822 & 0.648 & 0.521 & 0.558 & 0.534 & 0.508 & 0.508 & 0.537 & 0.523 & 0.332 & 0.294 & 0.712 & 0.713 \\
    BIECON ~\cite{BIECON} & 0.961 & 0.958 & 0.823 & 0.815 & 0.762 & 0.717 & 0.648 & 0.623 & 0.613 & 0.613 & 0.654 & 0.651 & 0.428 & 0.407 & {-} & {-} \\
    MEON ~\cite{MEON} & 0.955 & 0.951 & 0.864 & 0.852 & 0.824 & 0.808 & 0.691 & 0.604 & 0.710  & 0.697 & 0.628 & 0.611 & 0.394 & 0.365 & {-} & {-} \\
    WaDIQaM ~\cite{bosse2017deep} & 0.955 & 0.960  & 0.844 & 0.852 & 0.855 & 0.835 & 0.752 & 0.739 & 0.671 & 0.682 & 0.807 & 0.804 & 0.467 & 0.455 & {-} & {-} \\
    DBCNN ~\cite{zhang2018blind} & 0.971 & 0.968 & \underline{0.959} & \underline{0.946} & 0.865 & 0.816 & 0.856 & 0.851 & 0.869 & 0.851 & 0.884 & 0.875 & 0.551 & 0.545 & 0.915 & 0.911 \\
    TIQA ~\cite{TIQA} & 0.965 & 0.949 & 0.838 & 0.825 & 0.858 & 0.846 & 0.855 & 0.85  & 0.861 & 0.845 & 0.903 & 0.892 & 0.581 & 0.541 & {-} & {-} \\
    MetaIQA ~\cite{zhu2020metaiqa} & 0.959 & 0.960  & 0.908 & 0.899 & 0.868 & 0.856 & 0.775 & 0.762 & 0.802 & 0.835 & 0.856 & 0.887 & 0.507 & 0.54  & {-} & {-} \\
    P2P-BM ~\cite{ying2020patches} & 0.958 & 0.959 & 0.902 & 0.899 & 0.856 & 0.862 & 0.849 & 0.84  & 0.842 & 0.844 & 0.885 & 0.872 & 0.598 & 0.526 & {-} & {-} \\
    HyperIQA ~\cite{hypernet} & 0.966 & 0.962 & 0.942 & 0.923 & 0.858 & 0.840  & 0.845 & 0.852 & 0.882 & 0.859 & 0.917 & 0.906 & 0.602 & 0.544 & 0.915 & 0.911 \\
    TReS  ~\cite{TReS} & 0.968 & 0.969 & 0.942 & 0.922 & 0.883 & 0.863 & 0.858 & 0.859 & 0.877 & 0.846 & \underline{0.928} & 0.915 & 0.625 & 0.554 & {-} & {-} \\
    MUSIQ ~\cite{musiq} & 0.911 & 0.940  & 0.893 & 0.871 & 0.815 & 0.773 & 0.872 & 0.875 & 0.746 & 0.702 & \underline{0.928} & \underline{0.916} & \underline{0.661} & \underline{0.566} & \underline{0.921} & \underline{0.918} \\
    DACNN ~\cite{pan2022dacnn} & \underline{0.980}  & \underline{0.978} & {0.957} & {0.943} & \underline{0.889} & \underline{0.871} & \underline{0.905} & \underline{0.905} & \underline{0.884} & \underline{0.866} & 0.912 & 0.901 & {-} & {-} & \underline{0.921} & 0.915 \\
    %DEIQT ~\cite{DEIQT} & \textbf{0.982} & \textbf{0.980} & \textbf{0.963} & \textbf{0.946} & \textbf{0.908} & \textbf{0.892} & 0.887 & 0.889 & \textbf{0.894} & \textbf{0.875} & \textbf{0.934} & \textbf{0.921} & \textbf{0.663} & \textbf{0.571} & \textbf{0.923} & \textbf{0.919} \\
    \midrule
    \rowcolor{gray!25} RKIQT (ours) & \textbf{0.986} & \textbf{0.984} & \textbf{0.970} & \textbf{0.958} & \textbf{0.917} & \textbf{0.900} & \textbf{0.911} & \textbf{0.911} & \textbf{0.917} & \textbf{0.897} & \textbf{0.943} & \textbf{0.929} & \textbf{0.686} & \textbf{0.589} & \textbf{0.928} & \textbf{0.923} \\
    \bottomrule
    \end{tabular}}
  \caption{Performance comparison measured by averages of SRCC and PLCC, where bold entries indicate the best results, \underline{underlines} indicate the second-best.} 
  \label{performance}
\end{table*}
\end{center}
\begin{figure*}[htbp]
  \centering
    \includegraphics[width=1\textwidth,height=29mm]{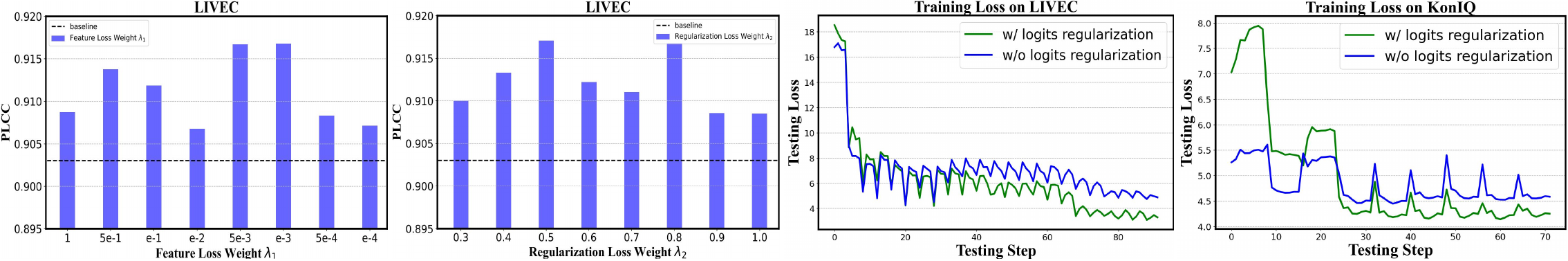}
  \caption{(a) and (b) are sensitivity experiment of hyper-parameters $\lambda_1$ and $\lambda_2$, Fig.~\ref{hyper_loss}(c) and (d) compare testing loss plots with regularization and baseline on LIVEC and KonIQ dataset which demonstrate the effectiveness of preventing overfitting}
  \vspace{-6pt}
  \label{hyper_loss}
\end{figure*}
\section{Experiments}
\subsection{Datasets and Evaluation Criteria}
We evaluate the performance of the proposed RKIQT on eight typical datasets, including four synthetic datasets of LIVE~\cite{sheikh2006statistical}, CSIQ~\cite{larson2010most}, TID2013~\cite{ponomarenko2015image}, KADID~\cite{lin2019kadid}, and four authentic datasets of LIVEC~\cite{ghadiyaram2015massive}
KonIQ~\cite{hosu2020KonIQ}, LIVEFB ~\cite{ying2020patches}, SPAQ ~\cite{fang2020perceptual}. Specifically, for the authentic dataset, LIVEC consists of 1162 images taken by different photographers using various mobile devices. SPAQ contains 11,125 photos collected from 66 smartphones. KonIQ-10k consists of 10073 images selected from public multimedia resources. LIVEFB is the largest real-world dataset to date and includes 39,810 images. For synthetic datasets, they contain a small number of original images that are synthetically distorted by various distortion types, such as JPEG compression and Gaussian blur. LIVE and CISQ contain 779 and 866 synthetic distorted images with five and six distortion types, respectively. TID2013 and KADID consist of 3000 and 10125 synthetically distorted images with 24 and 25 distortion types, respectively.
In our experiments, two commonly used criteria, Spearman's Order correlation coefficient (SRCC) and Pearson's linear correlation coefficient (PLCC), are used to quantify the performance of the model in terms of prediction monotonicity and prediction accuracy, respectively. The values of SRCC and PLCC range from -1 to 1. Superior performance should result in absolute values close to one for SRCC and PLCC.
\subsection{Implementation Details}
To train the student network, we follow the typical strategy of randomly cropping the input image into 10 image patches with a resolution of $224 \times 224$. Each image patch is then reshaped as a sequence of patches with a patch size of p = 16 and a dimension of input tokens as in $D = 384$. We create the Transformer encoder based on the ViT-S proposed in DeiT III \cite{touvron2022deit}, with the encoder depth set to 12 and the number of heads h = 6. The depth of the decoder is set to 1. The model is trained for 9 epochs with a learning rate of ${2 \times 10^{-4}}$ and a decay factor of 10 every 3 epochs. The batch size varies depending on the size of the dataset, with a batch size of 16 for LIVEC and 128 for KonIQ. For each dataset, 80\% of the images are used for training, and the remaining 20\% are used for testing. We repeat this process 10 times to mitigate performance bias and report the average of SRCC and PLCC.
For our CNN, INN teacher, and NAR-teacher, the pre-training setting follows a similar approach to the student training, and hyperparameter settings follow previous work~\cite{DEIQT}. Among them, CNN and INN teachers are pre-trained on 8 datasets, respectively. For NAR-teacher, pre-training is only performed on the synthetic dataset KADID. The teacher then performs offline distillation during student training.
%on 8 datasets, no more fine-tuning and parameter changes.

\subsection{Comparison with SOTA NR-IQA Methods}
Table \ref{performance} presents the comparative performance of the proposed RKIQT and other classical or state-of-the-art NR-IQA methods, including hand-crafted feature-based methods such as ILNIQE ~\cite{ILNIQE} and BRISQUE ~\cite{BRISQUE}, as well as deep learning-based methods such as MUSIQ ~\cite{ke2021musiq} and MetaIQA ~\cite{zhu2020metaiqa}. The evaluation results obtained from 8 diverse datasets demonstrate that RKIQT outperforms all other methods across each dataset. Notably, these datasets consist of varying image content and distortion types, making it extremely challenging to maintain consistent high performance across all of them. The observed performance gain of RKIQT over other NR-IQA methods confirms its effectiveness and superiority in accurately characterizing image quality.

\subsection{Generalization Capability Validation}
% 我们通过跨数据集验证方法进一步评估RKIQT的泛化能力,其中NR-IQA模型在一个数据集上训练,然后在其他数据集上测试,没有任何微调或参数适应。表2报告了五个数据集上SRCC平均数的实验结果。正如观察到的,RKIQT在六个跨数据集中的五个上实现了最佳性能,其中,在  LIVEC真实数据集上取得了明显的性能提升,并在KonIQ数据集上达到了竞争性能。这些结果表明RKIQT具有较强的泛化能力。

We evaluate the generalization ability of RKIQT by employing a cross-dataset validation approach. In this approach, we train the NR-IQA model on one dataset and test it on others without fine-tuning or parameter adaptation. Table \ref{cross-dataset} shows the experimental results of SRCC averages on the five datasets. As observed, RKIQT achieves the best performance on five of the six cross-datasets. It clearly outperforms the other methods on the LIVEC dataset and shows competitive performance on the KonIQ dataset which strongly demonstrates the generalization ability of RKIQT.
\begin{table}[t]
\small
\setlength\tabcolsep{4pt}
  \centering
    \begin{tabular}{ccccccc}
    \toprule
    Training & \multicolumn{2}{c}{  LIVEFB } & \multicolumn{1}{c}{LIVEC} & \multicolumn{1}{c}{KonIQ} & \multicolumn{1}{c}{LIVE} & \multicolumn{1}{c}{CSIQ} \\
    \midrule
    Testing & \multicolumn{1}{c}{KonIQ} & \multicolumn{1}{c}{LIVEC} & \multicolumn{1}{c}{KonIQ} & \multicolumn{1}{c}{LIVEC} & \multicolumn{1}{c}{CSIQ} & \multicolumn{1}{c}{LIVE} \\
    \midrule
    DBCNN & 0.716 & 0.724 & 0.754 & 0.755 & 0.758 & 0.877 \\
    P2P-BM & 0.755 & 0.738 & {0.740}  & 0.770  & 0.712 & {-} \\
    HyperIQA & \underline{0.758} & 0.735 & \textbf{0.772} & 0.785 & 0.744 & \underline{0.926} \\
    TReS  & 0.713 & \underline{0.740}  & 0.733 & \underline{0.786} & \underline{0.761} & {-} \\
    % DEIQT & 0.733 & 0.781 & 0.744 & 0.794 & 0.781 & \textbf{0.932} \\
    \midrule
    \rowcolor{gray!25} RKIQT   & \textbf{0.759} & \textbf{0.797} & \underline{0.760}  & \textbf{0.818} & \textbf{0.793} & \textbf{0.932} \\
    \bottomrule
    \end{tabular}%
  \caption{SRCC on the cross datasets validation. The best results are highlighted in bold, second-best are \underline{underlined}.}
  \label{cross-dataset}%
\end{table}%

\begin{figure*}[htbp]
	\centering{\includegraphics[width=171mm,height=81mm]{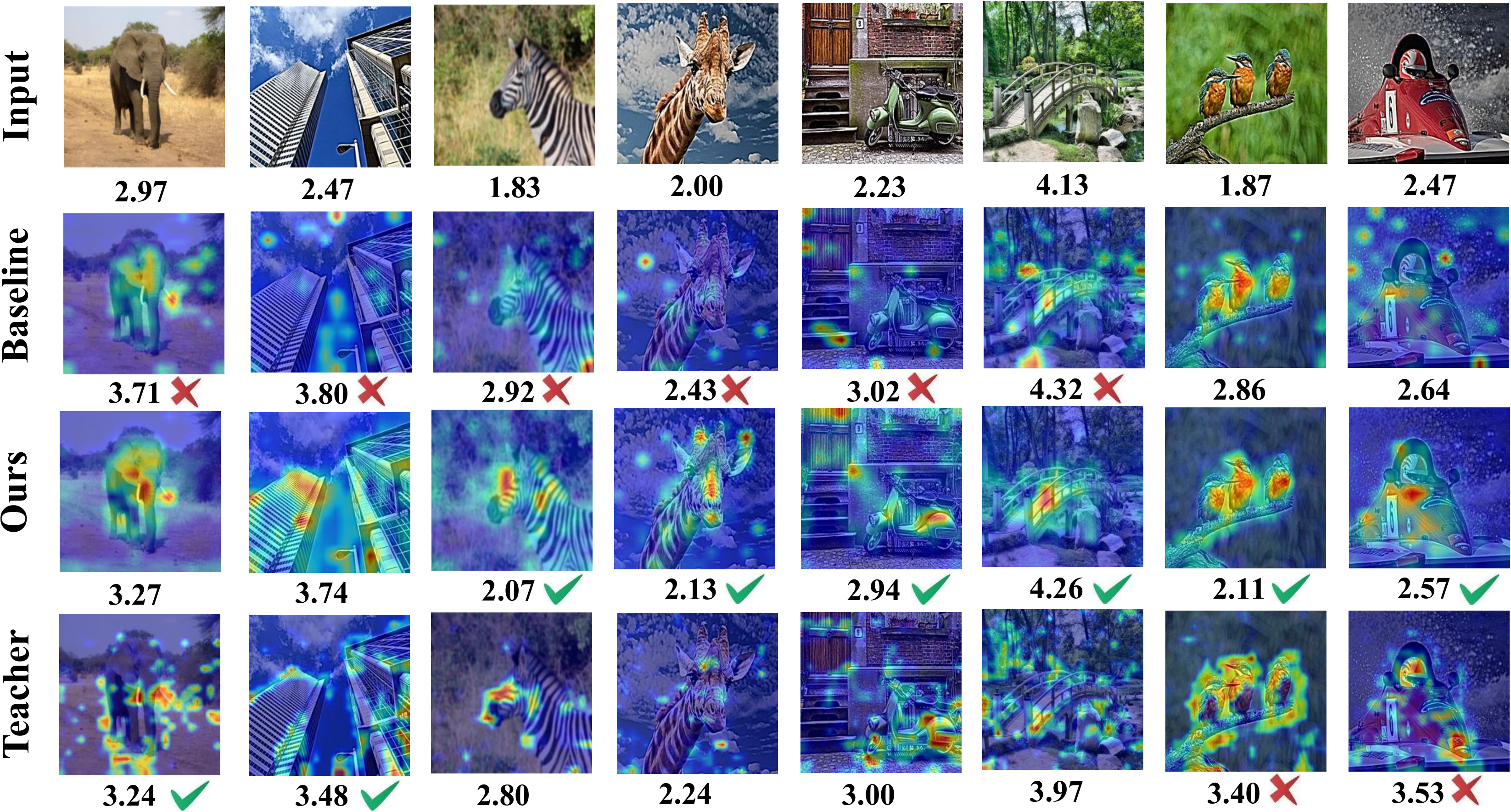}}
	\caption{Activation maps of baseline, RKIQT, and NAR-Teacher using the Grad-CAM~\cite{grad} method. Mean Opinion Scores (MOS) are displayed in the figures. 
 Our RKIQT model is designed to focus more on image distortion and consequently improves image quality prediction performance. Rows 1-4 represent input images, CAMs from baseline, RKIQT and NAR-Teacher. 
 Red crosses indicate the worst predictions, while green checkmarks indicate the best predictions.
 }
 \vspace{-10pt}
 \label{attention map}
\end{figure*}

% \begin{figure}[htbp]
% 	\centering{\includegraphics[width=0.45\textwidth]{figure/热力图.pdf}}
% 	\caption{Cosine similarity between perceptual features of CLS token, CNN token, and INN token. The low similarity between CNN/INN token and CLS token suggests that each token judges the image quality from a unique  perspective.
%  }
%  \label{heatmap}
% \end{figure}

\subsection{Ablation Study}
\textbf{Ablation on overall Distillation framework.}
% RKIQT由三个基本模块组成,包括NR student、Feature蒸馏Teacher和Logits蒸馏Teacher。我们进行消融实验,以检查每个模块的个人贡献。表3显示了LIVEC和KonIQ数据集上的实验结果。表3中的NR指的是输入只有失真图,结构由ViT编码器、transformer解码器构成的encoder-decoder框架。w/o Inductive表示没有Logits阶段的蒸馏。w/o NAR表示没有Feature阶段的蒸馏。
RKIQT consists of Masked Quality-Contrastive Distillation (MCD) and Inductive Bias Regularization. We perform ablation studies to examine the individual contribution of each module. Table~\ref{abaltion_overall} shows the experimental results on KADID, LIVEC, and KonIQ datasets. The w/o Regular. denotes feature distillation without the Inductive Bias Regularization, and w/o NAR represents regular without Masked Quality-Contrastive Distillation.

From Table \ref{abaltion_overall}, we observe that both MCD and Inductive Bias Regularization are very effective in characterizing image quality, thus contributing to the overall performance of RKIQT. Surprisingly, according to our distillation learning framework, this allows our model to \textbf{outperform} our Teacher which has reference Prior information in the case of distillation (The gray part in Table \ref{abaltion_overall}).
To be specific, the proposed inductive bias regularization approach significantly enhances the accuracy and stability of the model, while our MCD technique has a more significant impact on the KADID dataset. This outcome is reasonable since the inductive bias regularization involves a more expensive pre-training cost, where each dataset is pre-trained with the corresponding teacher, thereby introducing significantly more prior information than what is introduced by MCD. However, MCD still enables our model to achieve better performance and generalization than existing state-of-the-art NR-IQA methods with less pretraining. 
In conclusion, the ablation study demonstrates the significant contribution of each component to the RKIQT. The proposed MCD and regularization provide notable improvements in model accuracy and  stability, and their combination leads to a sota model for image quality assessment. More detailed ablation experiments on MCD, Inductive Bias Regularization, and Inductive Bias token can be obtained in the supplementary material.
\begin{table}[t]
\small
\setlength\tabcolsep{3.5pt}
  \centering
    \begin{tabular}{lcccccc}
    \toprule
    & \multicolumn{2}{c}{KADID} & \multicolumn{2}{c}{LIVEC} & \multicolumn{2}{c}{KonIQ} \\
    \cmidrule{2-7}    Method & \multicolumn{1}{c}{PLCC} & \multicolumn{1}{c}{SRCC} & \multicolumn{1}{c}{PLCC} & \multicolumn{1}{c}{SRCC} & \multicolumn{1}{c}{PLCC} & \multicolumn{1}{c}{SRCC} \\
    \midrule
    {} & {} & {} & \textbf{Teacher} & {} & {}\\
    CNN-teacher & 0.865 & 0.866 & {0.892} & {0.866} & {0.921} & {0.903} \\
    INN-teacher & 0.789 & 0.798 & {0.815} & {0.811} & {0.910} & {0.900} \\
    \rowcolor{gray!25} NAR-teacher &  0.909 & 0.902 & {-} & {-} & {-} & {-} \\
    \midrule
    {} & {} & {} & \textbf{Student} & {} & {}\\
    baseline & 0.878 & 0.884 & 0.887 & 0.865 & 0.930 & 0.918 \\
    \mbox{w/o Regular.}  & 0.903 & 0.905 & 0.903 & 0.879 & 0.938 & 0.927 \\
    \mbox{w/o NAR}  & 0.902 & 0.902 & 0.907 & 0.881 & 0.939 & 0.926 \\
    \rowcolor{gray!25} RKIQT & {\textbf{0.911}} & {\textbf{0.911}} & \textbf{0.917 } & \textbf{0.897} & \textbf{0.943} & \textbf{0.929} \\
    \bottomrule
    \end{tabular}%
  \caption{Ablation experiments on KADID, LIVEC, and KonIQ datasets. Bold entries indicate the best performance. RKIQT outperforms the NAR-teacher when all modules are employed.}
  \label{abaltion_overall}%
  \vspace{-5pt}
\end{table}%

\textbf{Ablation on Inductive Bias Regularization.}
The proposed inductive bias regularization has two main effects: (1) preventing overfitting and (2) accelerating convergence. 
To demonstrate the overfitting, we compare the test loss during training with the baseline test loss, as shown in Fig. \ref{hyper_loss}. The results show that, at the end of training on both datasets, the regularization achieves a lower test error than the baseline. As the test step size reaches 70/50, the baseline exhibits larger test errors and larger oscillations, with significant overfitting. In contrast, the regularization shows a steady decrease in the test error. It's important to note that during the early training phase on the LIVEC dataset (before 35 steps), the impact of regularization was somewhat limited. However, after reaching 70 steps, the benefits of regularization became much more pronounced. This change can be attributed to the initially poor performance of CNN and INN teachers on the LIVEC dataset, as indicated in Table~\ref{abaltion_overall}. This poor performance may have initially led to negative regularization effects. But as training progressed, our regularization approach gradually reduced the gap between the teacher and student models through reverse distillation. Consequently, we observed a significant improvement after extending the training to 70 steps. Therefore, we conclude that with adequate training, the proposed logit regularization is effective in preventing overfitting. Regarding accelerating convergence, please refer to Supplementary Material.

\section{Qualitative Analysis}~\label{sec:attentionmap}
\textbf{Analysis on Sensitivity of hyper-parameters.}
In this paper, we use $\lambda_1$ and $\lambda_2$ in Eq.~\ref{loss equ} to balance the MCD and regularization, respectively. In this subsection, we do the sensitivity study of the hyperparameters and conduct experiments on different Loss weights $\lambda$ to explore their effect on RKIQT. As shown in Fig.~\ref{hyper_loss}, the MCD and Inductive Bias Regularization is not very sensitive to the hyper-parameter $\lambda$, which is just used for balancing the loss. This indicates that the choice of hyper-parameter in our approach is relatively arbitrary, highlighting the robustness of our model.

\textbf{Visualization of quality attention map.}
In this study, we utilize GradCAM \cite{grad} to generate a visual representation of the feature attention map, as shown in Fig. \ref{attention map}. We observe that the teacher model primarily focuses on global edges rather than semantic information, suggesting that teachers focus on edges as an important characteristic of image quality. Our proposed RKIQT, on the other hand, accurately and comprehensively focuses on the target distortion region, leveraging the learned prior information of high-quality images. In contrast, the baseline model is more prone to distractions and may even focus on non-distortion regions. We find that the baseline model is less effective at assessing the quality with clear foreground objects. However, RKIQT benefits from the contrastive awareness learned from the teacher model, enabling it to learn the key elements of difference information between HQ and LQ images, effectively extracting quality perception features.

The predicted results indicate that RKIQT performs better than the baseline across different distortion levels, and in most cases, it outperforms the teacher model as well. However, the two leftmost columns show some failed cases. For the failed, limited by without reliable images reference which leads to the indistinguishability to images with severe edge distortion, Neither RKIQT nor the baseline model performs well in such cases. Nonetheless, RKIQT still identifies appropriate regions more accurately than the baseline model and selects the correct distortion region.

\section{Conclusion}
The primary challenge for NR-IQA is the absence of effective reference information. To mitigate this issue, we introduce the reference knowledge into the NR-IQA and propose an RKIQT method. We make the first attempt to introduce human comparative thinking into the IQA model, thus ensuring a high consistency with the human subjective evaluation. In particular, we design a Masked Quality-Contrastive Distillation module that distills teachers' comparison knowledge given non-aligned high-quality images. Students learn such knowledge through partial feature patches, which render a high robustness and feature extraction ability. Furthermore, an inductive bias regularization is proposed based on the CNN and INN networks. It allows the students with fewer inductive biases to learn from teachers with various inductive biases, and subsequently achieve a high rate of convergence and generalization capability. Experiments on 8 benchmark IQA datasets verify the superiority of the RKIQT.
{
\bibliography{aaai24}
}
\medskip
\clearpage
\appendix
\section{Appendix Overview}
The supplementary material is organized as follows:
\textbf{Explanations for Concepts:} provide explanations for some of the concepts in the manuscript.
\textbf{Training and Evaluation Details:} shows more training and evaluation details. 
\textbf{More ablation:} provides more ablation experiments, including MCD, random mask, Inductive Bias token, and Inter-layer.
\textbf{Related Work:} provides more discussion about KD.
\section{Explanations for Concepts}
\noindent\textbf{ClS Token:} In classification tasks, the input image is divided into multiple patches.  The Vision Transformer (VIT) model learns to extract global aggregate information by aggregating relationships between different patches through learnable CLS tokens.

\noindent\textbf{Token Inductive Bias:} This bias assigns different biases to tokens.  The purpose is to align the inductive bias of tokens with that of teachers, enabling tokens to learn more effectively from their corresponding teachers.

\noindent\textbf{Expansion of INN:} Involution (INN) is a type of kernel that is shared across channels but distinct in spatial extent.  INN exhibits precisely inverse inherent characteristics compared to convolution, enabling it to capture global spatial relationships in an image.

\noindent\textbf{Pixel-Aligned Reference:} This term refers to a clear reference image that corresponds to a distorted image, having exactly the same content information as the distorted image.

\noindent\textbf{Offline Distillation:} During training, knowledge from a pre-trained teacher model is transferred to a student network. Only the student network is trained, while the parameters of the teacher network are frozen.

\noindent\textbf{Non-aligned Reference:}
In this paper, "aligned" refers to situations where a blurred image has a corresponding clear image of the same version. For instance, if we have a blurry photo due to camera shake, two images are considered aligned when the camera captures the distorted image's corresponding clear image under the same scene, view angle, and lighting conditions. However, obtaining this aligned clear image is often challenging in practical settings. Typically, the reference image used is non-aligned. Therefore, the term "non-aligned model" means that the pixel of low-quality image and the high-quality image don't have a one-to-one correspondence. In other words, the high-quality image only needs to be clear, while the image content can vary.
\begin{algorithm}[H]
\caption{Inference Process of NR-Student}
\label{alg:testing}
\begin{algorithmic}[1]
\REQUIRE
\STATE Low-quality (LQ) images: $X_{LQ}$
\STATE Inductive Bias Student: $S$

\STATE \textbf{Testing Process:}
\STATE Using ClS token, Conv token, and Inv token in the $S$ to get the quality score $Y_{cls}$, $Y_{s_{cnn}}$, $Y_{s_{inn}}$.
\STATE Select $Y_{cls}$ as the final output.

\textbf{Output:} $Y_{cls}$
\end{algorithmic}
\end{algorithm}
\begin{algorithm}[H]
\caption{Training Process of NR-Student}
\label{alg:training}
\begin{algorithmic}[1]
\REQUIRE
\STATE Low-quality (LQ) images: $X_{LQ}$
\STATE LQ images' ground truth: $Y_{gt}$
\STATE High-quality (HQ) images: $X_{HQ}$
\STATE Inductive Bias Student Network: $S$
\STATE CNN teacher's learnable intermediate layer: $T^l_{cnn}$
\STATE INN teacher's learnable intermediate layer: $T^l_{inn}$
\STATE Encoder layer number $i$, $1 \leq i \leq L$
\STATE Non-aligned reference teacher (NAR-teacher): $T_{nar}$
\STATE Pre-trained CNN teacher: $T_{cnn}$, INN teacher: $T_{inn}$
\STATE Loss hyper-parameters: $\lambda_1$, $\lambda_2$

\STATE \textbf{Masked Quality-Contrastive Distillation:}
\FOR{each encoder layer $i = 1,2,...,L$}
    \STATE Obtain $F_{LQ}$ of input $X_{LQ}$ using the $S$ encoder.
    \STATE Obtain LQ-HQ difference-aware features $F_{HQ-LQ}$ using the $T_{nar}$ encoder.
    \STATE Randomly mask $F_{LQ}$ to obtain $F_{mask}$.
    \STATE Generation module to restore $F_{mask}$ to the $F_{new}$.
    \STATE MSE loss between $F_{new}$ and $F_{HQ-LQ}$: $\mathcal{L}_{fea}^i$.
\ENDFOR
\STATE Sum up $\mathcal{L}_{fea}^i$ of all layers.
\STATE \textbf{Inductive Bias Regularization:}
\STATE Get the output $Y_{cls}$, $Y_{s_{cnn}}$, $Y_{s_{inn}}$ using the $S$.
\STATE Obtain ${Y_{{T}'_{cnn}}}$, $Y_{{T}'_{inn}}$ of input $X_{LQ}$ using $T_{cnn}$ and $T_{inn}$.
\STATE Obtain pseudo-label $Y_{T_{cnn}}$, $Y_{T_{inn}}$ of input $X_{LQ}$ using $T^l_{cnn}$ and $T^l_{inn}$, respectively.
\STATE Calculate loss $\mathcal{L}_{logits}$ in Equ.~3,5 of our manuscript.
\STATE Calculate loss $\mathcal{L}_{all}$ in Equ. 6 of our manuscript.
\STATE Use $\mathcal{L}_{all}$ to update $S$.

\textbf{Output:} $S$
\end{algorithmic}
\end{algorithm}
\section{Training and Evaluation Details}
Our teacher models are both pre-trained and freeze parameters during student training. 

\noindent\textbf{Training Stage:} As depicted in Fig.~\ref{modelStructure} and algorithm~\ref{alg:training}, begins with an input image.  The student model, along with three different inductive bias tokens, and the NAR-teacher model, acquire both LQ features and the difference in distribution between HQ and LQ features.  To improve the student's feature representation, we employ Mask Quality Contrast distillation.  This involves masking the student's feature map and generating a new feature using a simple generation module.  The generation process is supervised by the NAR-teacher's differential features.   Subsequently, the student's three different inductive bias tokens enter the decoder to predict three quality scores.  Each quality score is supervised by a specific inductively biased teacher.  However, instead of directly using the teacher logits with different inductive biases to supervise the students, we introduce a learnable intermediate layer.  This is done to mitigate the potential large quality perception gap between teachers and students. Additionally, it is worth noting that the learnable intermediate layer is supervised by both the students and the CNN and INN teachers.

\noindent\textbf{Inference Stage:} All teacher models, feature distillation, and regularization techniques are no longer utilized. In other words, as depicted in algorithm~\ref{alg:testing}, the student model is directly applied for inference without reference images or high-quality images.

\section{More ablation}~\label{ablation_mcd}
\textbf{Ablation on Masked Quality-Contrastive Distillation.}
\begin{table}[t]
\small
\setlength\tabcolsep{4pt}
  \centering
    \begin{tabular}{lcccccc}
    \toprule
    & \multicolumn{2}{c}{LIVE} & \multicolumn{2}{c}{LIVEC} & \multicolumn{2}{c}{KonIQ} \\
    \cmidrule{2-7}    Method & \multicolumn{1}{c}{PLCC} & \multicolumn{1}{c}{SRCC} & \multicolumn{1}{c}{PLCC} & \multicolumn{1}{c}{SRCC} & \multicolumn{1}{c}{PLCC} & \multicolumn{1}{c}{SRCC} \\
    \midrule
    baseline   & 0.978 & 0.977 & 0.887 & 0.865 & 0.930 & 0.918 \\
    std   & ±0.004 & ±0.005 & ±0.02 & ±0.017 & ±0.003 & ±0.004 \\
    w/ DRD   & 0.983 & 0.981 & 0.908 & 0.889 & 0.940  & 0.925 \\
    std   & ±0.003 & ±0.003 & ±0.008 & ±0.014 & ±0.002 & ±0.003 \\
    \rowcolor{gray!10} w/ MCD   &  \textbf{0.986} & \textbf{0.984} & \textbf{0.917} & \textbf{0.897} & \textbf{0.943} & \textbf{0.929} \\
    \rowcolor{gray!10} std   & \textbf{±0.002} & \textbf{±0.003} & \textbf{±0.008} & \textbf{±0.009} & \textbf{±0.002} & \textbf{±0.002} \\
    % Teacher & \textbf{0.986} & \textbf{0.984} & \textbf{0.917} & \textbf{0.894} & \textbf{0.943} & \textbf{0.928} \\
    \bottomrule
    \end{tabular}%
  \caption{MCD ablation experiments on LIVE, LIVEC, and KonIQ datasets. Bold entries indicate the best performance.}   
  \label{ablation_distill}
\end{table}%
To further investigate the effectiveness of the proposed MCD, we conduct ablation experiments to train the model by changing the way of feature distillation to MCD and direct feature distillation (DRD), respectively. We repeat the experiment 10 times for each set of training data and report the average of PLCC, and SRCC.
The experimental results are detailed in Table \ref{ablation_distill}. Training model via MCD achieves the best accuracy compared to DRD on both synthetic and authentic datasets, especially on the authentic dataset LIVEC
These observations vividly show that the distillation way of MCD enhances the robustness of the model to image distortion perception in natural environments. In other words, RKIQT effectively utilizes the information of the asymmetric reference graph and achieves the best performance on both synthetic and real datasets.
\begin{figure}[t]
    \centering
    \includegraphics[width=0.46\textwidth]{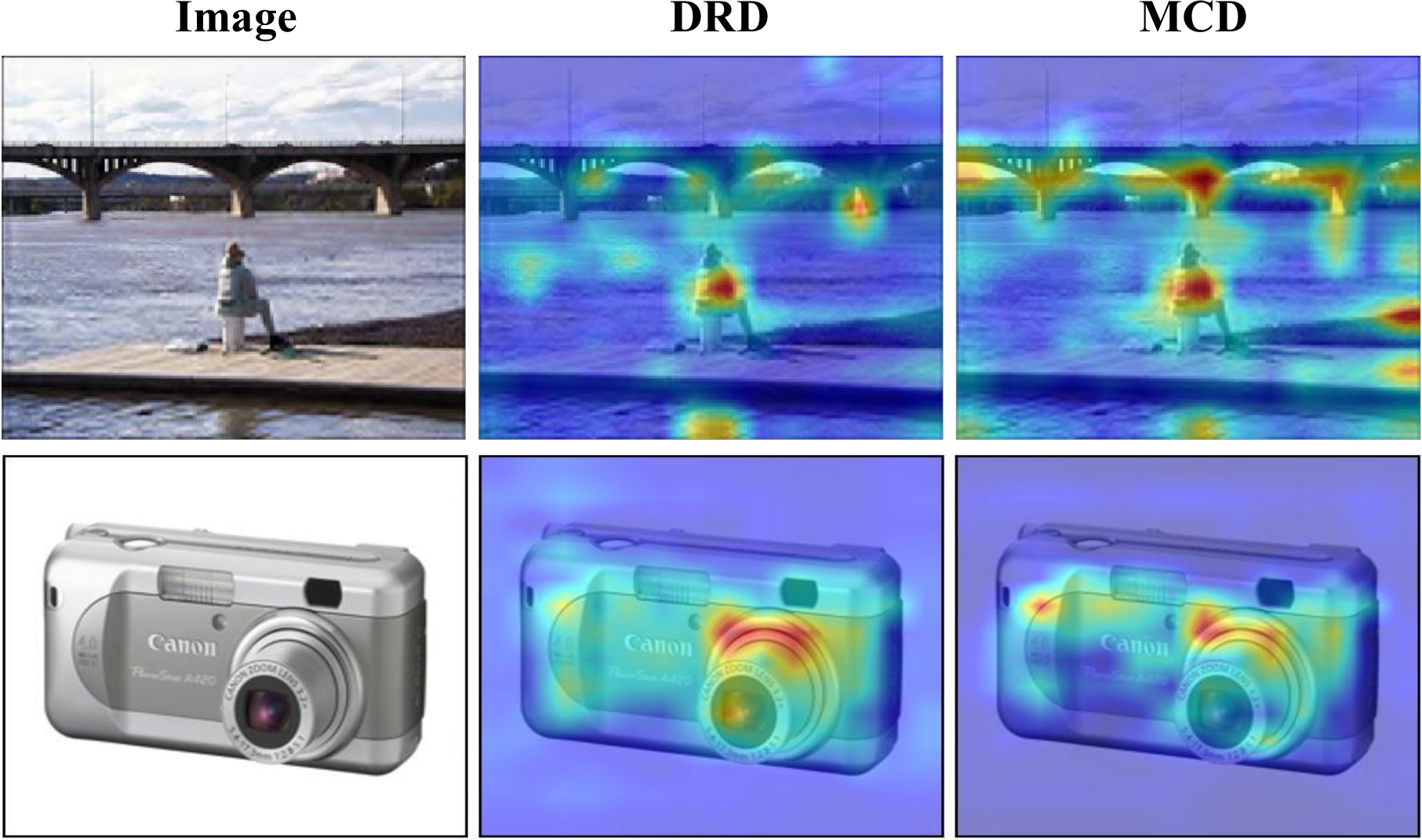}
    \caption{ Images from left to right are the input images and their attention maps with DRD and MCD. As observed, MCD pays more attention to the background distortion region, and to the quality distortion region of the subject}
    \label{MCD}
\end{figure}
% 表征能力强
\begin{figure}[t]
    \centering
    \includegraphics[width=0.46\textwidth]{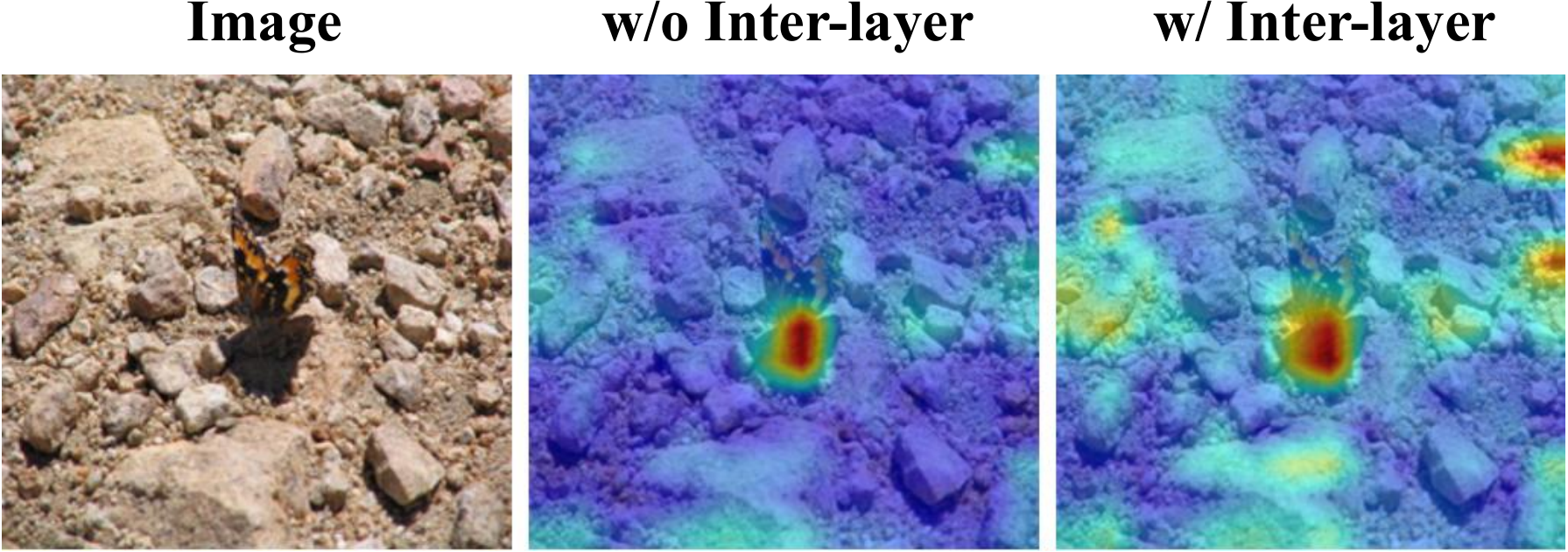}
    \caption{The first picture is the distorted picture. The remaining images are the attention map without and with learnable Inter-layer, respectively. Incorporating the Inter-layer, our model pays more attention to the quality-aware features.}
    \label{shadow}
\end{figure}
We provide a detailed analysis and consider that (i) MCD aids the model in acquiring HQ-LQ distribution difference knowledge (i.e., contrastive ideas) and (ii) MCD preserves both local distortion and global semantic features in the masked pixels, in conjunction with (i) to generate more comprehensive quality-aware features. 
It is important to note that HQ-LQ distribution difference knowledge is mainly represented by the edge of foreground and background in visualization, as illustrated in row 4 of Fig.~\ref{attention map} of the manuscript. This is further demonstrated in Fig.\ref{MCD}, which presents images containing complex content (top) and simple content (bottom), accompanied by the corresponding student encoder visualization outcomes. When the image is relatively simple, MCD's response to background quality perception is significantly reduced, with greater focus placed on the distortion of the foreground content, thus confirming the second point (ii). However, as the complexity of the image scene increases, MCD also starts to respond more to the quality perception of the edge background, thus supporting the first point (i).
\begin{figure}[t]
	\centering{\includegraphics[width=0.45\textwidth]{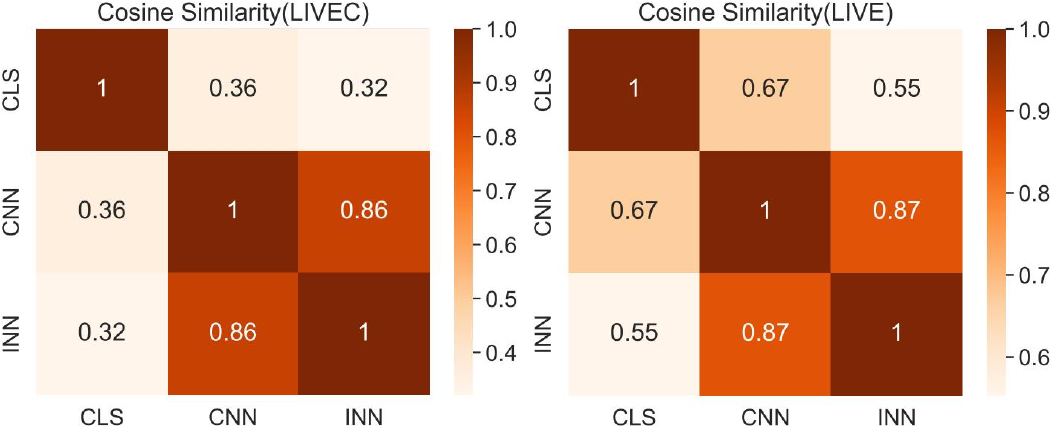}}
	\caption{Cosine similarity between perceptual features of CLS token, CNN token, and INN token. The low similarity between CNN/INN token and CLS token suggests that each token judges the image quality from a unique  perspective.
 }
 \label{heatmap}
\end{figure}

\begin{figure}[t]
\vspace{-0.2cm}
    \centering
\includegraphics[width=0.47 \textwidth]{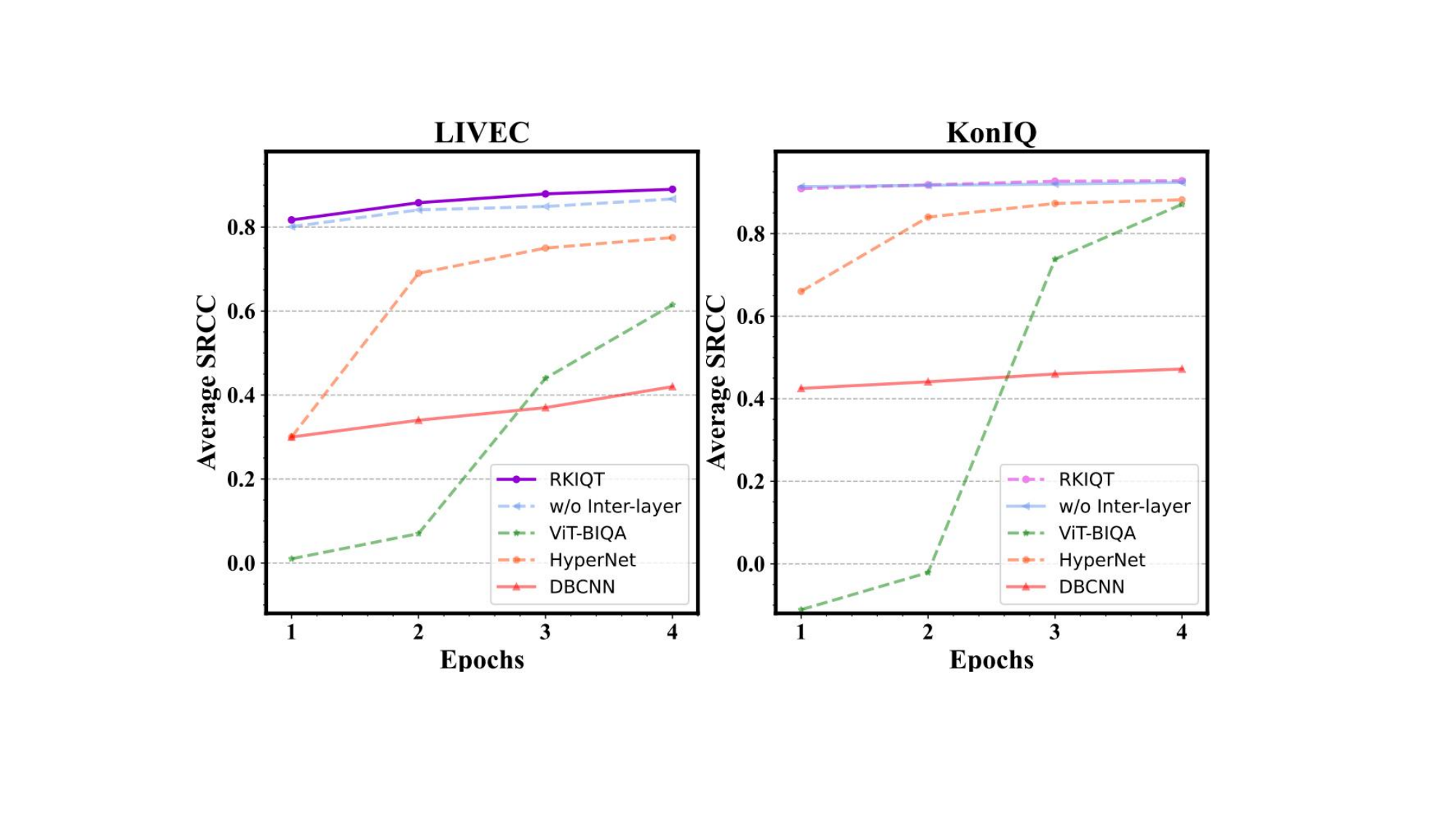}
    \caption{Average SRCC versus Epochs on different datasets ablation on Inductive Bias Regularization }
    \label{efficient}
  \vspace{-0.2cm}
\end{figure}

\noindent\textbf{The effectiveness of accelerating convergence.} To demonstrate the effect of the regularization on convergence, we evaluate the training efficiency and performance of RKIQT distillation, as shown in Fig. \ref{efficient}, which depicts the SRCC with an increasing number of epochs on LIVEC and KonIQ test sets. The results show that RKIQT converges significantly faster than the other methods, achieving the fastest convergence after only one epoch of training, which outperforms the second-best NR-IQA method in Table~\ref{performance} of the manuscript. Furthermore, on LIVEC, the use of the Inter-layer module greatly reduces the negative impact of the teacher network's less ideal performance, indicating that the Inter-layer module preserves the diversity of knowledge and accelerates convergence. These observations demonstrate that RKIQT and the teacher can "learn from each other", with the teacher adapting its teaching to the student's abilities, resulting in more comprehensive knowledge and significantly improved model stability.

\textbf{Ablation on Inductive Bias token.}
To demonstrate that these tokens with different inductive biases indeed model unique features, we compute the cosine similarity between the CLS, CNN, and INN tokens of the distillation model (results are averaged over the LIVEC and LIVE datasets, respectively). as shown in Fig.~\ref{heatmap}, the result is between 0.32 and 0.7. This is significantly lower than the similarity between class and distillation labels in previous work~\cite{touvron2021training}; 0.96 and 0.94 in DeiT-T and Deit-S, respectively. This confirms our hypothesis that modeling local and global features with multiple perspectives separately with separate tokens in Vits leads to a more comprehensive quality-aware feature representation.

\textbf{Ablation on Inter-layer in Inductive Bias Regularization.}
To better understand the effectiveness of the learnable intermediate layer, we conduct additional experiments, as shown in Table \ref{tab:inter layer}.  Although our regularization, even without inter-layer still outperforms SOTA NR-IQA in Table 1, there is still considerable scope for enhancing the stability of the model. Adding Inter-layer modules can further enhance the model's performance. In addition to its strong regularization ability, we confirm its contribution to the texture extraction ability of the model, as shown in Fig. \ref{shadow}. The model can accurately perceive low contrast and low detail (such as stones) in the image. These observations demonstrate that the Inter-layer strategy effectively enables the model to learn prior knowledge and texture information from various bias NR-IQA models, thereby significantly improving the training efficiency.
\begin{table}[t]
\small
\setlength\tabcolsep{8.5pt}
  \centering
    \begin{tabular}{lccccc}
    \toprule
    & \multicolumn{2}{c}{LIVEC} & \multicolumn{2}{c}{KonIQ} \\
    \cmidrule{2-5}    Method & \multicolumn{1}{c}{PLCC} & \multicolumn{1}{c}{SRCC} & \multicolumn{1}{c}{PLCC} & \multicolumn{1}{c}{SRCC} \\
    \midrule
    % {} & {}  & \textbf${Teacher}$ & {} & {}\\
    % CNN-teacher & 0.892 & 0.866 & 0.921 & 0.903 \\
    % INN-teacher & 0.815 & 0.812 & 0.910 & 0.900 \\
    % \midrule
    % {} & {}  & \textbf${Student}$ & {} & {}\\
    baseline   & 0.894 & 0.875 & 0.935 & 0.922 \\
    std   & ±0.02 & ±0.017 & ±0.003 & ±0.004 \\
    w/o Inter-layer   & 0.911 & 0.886 & 0.941  & 0.928 \\
    std   & ±0.009 & ±0.014 & ±0.004 & ±0.003 \\
    \rowcolor{gray!10} w/ Inter-layer  & \textbf{0.917} & \textbf{0.897} & \textbf{0.943} & \textbf{0.929} \\
    \rowcolor{gray!10} std   & \textbf{±0.008} & \textbf{±0.009} & \textbf{±0.002} & \textbf{±0.002} \\
    \bottomrule
    \end{tabular}%
  \caption{Inter-layer ablation experiments on LIVEC and KonIQ datasets. Bold entries indicate the best performance.}  
  \label{tab:inter layer}
\end{table}%

\textbf{Ablation on Random Mask.}
\begin{table}[t]
\small
\setlength\tabcolsep{15pt}
  \centering
    \begin{tabular}{lcc}
\toprule
Method & PLCC & SRCC \\
\midrule
RKIQT w/ random mask & 0.917 & 0.897 \\
RKIQT w/ Gaussian(center) & 0.916 & 0.897 \\
RKIQT w/ all mask(center) & 0.916 & 0.896 \\
RKIQT w/ Gaussian(edge) & \textbf{0.919} & 0.896 \\
RKIQT w/ all mask(edge) & 0.918 & \textbf{0.900} \\
\bottomrule
\end{tabular}
\caption{Mask function ablation experiments were carried out on LIVEC datasets. Bold entries indicate the best performance.}
\label{mask}
\end{table}
Given that local distortions are often concentrated in the foreground or center regions of an image, we conducted four sets of experiments to investigate the effects of local distortion erasure, as shown in Table.~\ref{mask}. These experiments focused on the center and edge regions of the image.

\noindent{(1)RKIQT W/ Gaussian(center)}: The random mask function was replaced with a Gaussian distribution probability mask function, and the central region of the feature map was masked with a higher probability.

\noindent{(2)RKIQT W/ Gaussian(edge)}: The random mask function was replaced with a Gaussian distribution probability mask function, and the edge region of the feature map was masked with a higher probability.

\noindent{(3)RKIQT W/ all mask(center)}: In this experiment, all blocks in the central region were masked, while the edge region was masked with a lower probability.

\noindent{(4)RKIQT W/ all mask(edge)}: In this experiment, all blocks in the edge region were masked, while the central region was masked with a lower probability.

From the experimental results shown in Table~\ref{mask}, we conducted two sets of experiments to mask the central region.  Interestingly, the experimental results indicate that masking the central region had almost no impact on the performance of our model.  On the contrary, when we considered applying a larger probability of masking to the edge region or even masking the entire image except for the central region, we observed some improvement in the model's performance.  These findings suggest that the erasure of local distortions has little effect on the model's performance, and in some cases, an appropriate masking mechanism can even enhance the model's performance. This provides a potential direction for our future work.

\section{Related Work}
\noindent\textbf{Knowledge Distillation.}
Recent advancements in knowledge distillation have been significant. Hinton et al.~\cite{hinton2015distilling} laid the foundational concept of training a smaller 'student' model to replicate a larger 'teacher' model. Tung and Mori~\cite{tung2019similarity} innovated this by focusing on feature maps, and Cho and Hariharan~\cite{cho2019efficacy} introduced attention mechanisms into distillation processes. Mirzadeh et al.~\cite{mirzadeh2020improved} added the concept of an assistant network for effective distillation.
Some recent works in Image Quality Assessment (IQA) like Zheng et al. (2021)\cite{zheng2021learning} and Yin et al. (2022)\cite{yin2022content} have explored using KD to transfer reference information to student models. This approach aims to reduce the student models' dependency on the availability of reference images, leading to the development of degraded-reference IQA (DR-IQA) and non-aligned reference IQA (NAR-IQA) methods. However, these methods still face limitations due to their reliance on reference images, which is impractical for NR-IQA.
To the best of our knowledge, we make the first attempt to transfer more HQ-LQ difference prior information to the NR-IQA via KD, endowing students with the awareness of comparison. Experiments prove that distillation operations can further help our students achieve more accurate and stable performance.
% \bibliography{aaai24}
 
\end{document}